\documentclass[letterpaper]{article} % DO NOT CHANGE THIS
\usepackage{aaai25}  % DO NOT CHANGE THIS
\usepackage{times}  % DO NOT CHANGE THIS
\usepackage{helvet}  % DO NOT CHANGE THIS
\usepackage{courier}  % DO NOT CHANGE THIS
\usepackage[hyphens]{url}  % DO NOT CHANGE THIS
\usepackage{graphicx} % DO NOT CHANGE THIS
\usepackage{natbib}  % DO NOT CHANGE THIS AND DO NOT ADD ANY OPTIONS TO IT
\usepackage{caption} % DO NOT CHANGE THIS AND DO NOT ADD ANY OPTIONS TO IT
\usepackage{algorithm}
\usepackage{algorithmic}
\usepackage{algorithm}
\usepackage{algorithmic}
\usepackage{subcaption}
\usepackage{multirow}
\usepackage{tikz}

\usepackage{booktabs}
\usepackage{amsfonts}
\usepackage{amsmath}
\usepackage{amssymb}
\usepackage{graphicx}
\usepackage{bm}
\usepackage{appendix} % to handle appendices
\usepackage{nameref} % for referencing section names
\usepackage{float}
\usepackage{comment}

\definecolor{mygreen}{RGB}{84,130,53}
\definecolor{myred}{RGB}{192,0,0}

\definecolor{myyellow}{RGB}{252,249,234}

\usepackage{newfloat}
\usepackage{listings}
\DeclareCaptionStyle{ruled}{labelfont=normalfont,labelsep=colon,strut=off} % DO NOT CHANGE THIS
\floatstyle{ruled}
\newfloat{listing}{tb}{lst}{}
\floatname{listing}{Listing}
\title{Interweaving Memories of a Siamese Large Language Model}
\author {
    Xin Song\textsuperscript{\rm 1},
    Zhikai Xue\textsuperscript{\rm 2},
    Guoxiu He\textsuperscript{\rm 1}\thanks{Corresponding author},
    Jiawei Liu\textsuperscript{\rm 3},
    Wei Lu\textsuperscript{\rm 3}
}
\affiliations {
    \textsuperscript{\rm 1}School of Economics and Management, East China Normal University\\
    \textsuperscript{\rm 2}Department of Computer Science, Worcester Polytechnic Institute\\
    \textsuperscript{\rm 3}School of Information Management, Wuhan University\\
    xsong2023@stu.ecnu.edu.cn, 
    zxue1@wpi.edu, gxhe@fem.ecnu.edu.cn, \{laujames2017, weilu\}@whu.edu.cn

}

\begin{document}

\maketitle

\begin{abstract}
Parameter-efficient fine-tuning (PEFT) methods optimize large language models (LLMs) by modifying or introducing a small number of parameters to enhance alignment with downstream tasks. However, they can result in catastrophic forgetting, where LLMs prioritize new knowledge at the expense of comprehensive world knowledge. A promising approach to mitigate this issue is to recall prior memories based on the original knowledge. To this end, we propose a model-agnostic PEFT framework, \textbf{IMSM}, which \textbf{I}nterweaves \textbf{M}emories of a \textbf{S}iamese Large Language \textbf{M}odel. Specifically, our siamese LLM is equipped with an existing PEFT method. Given an incoming query, it generates two distinct memories based on the pre-trained and fine-tuned parameters. IMSM then incorporates an interweaving mechanism that regulates the contributions of both original and enhanced memories when generating the next token. This framework is theoretically applicable to all open-source LLMs and existing PEFT methods. We conduct extensive experiments across various benchmark datasets, evaluating the performance of popular open-source LLMs using the proposed IMSM, in comparison to both classical and leading PEFT methods. Our findings indicate that IMSM maintains comparable time and space efficiency to backbone PEFT methods while significantly improving performance and effectively mitigating catastrophic forgetting. 
\end{abstract}
\begin{links}
    \link{Code}{https://github.com/ECNU-Text-Computing/IMSM}
\end{links}

\section{Introduction}
\label{sec:intro}

\begin{figure}[h]
	\centering

\includegraphics[width=0.45\textwidth]{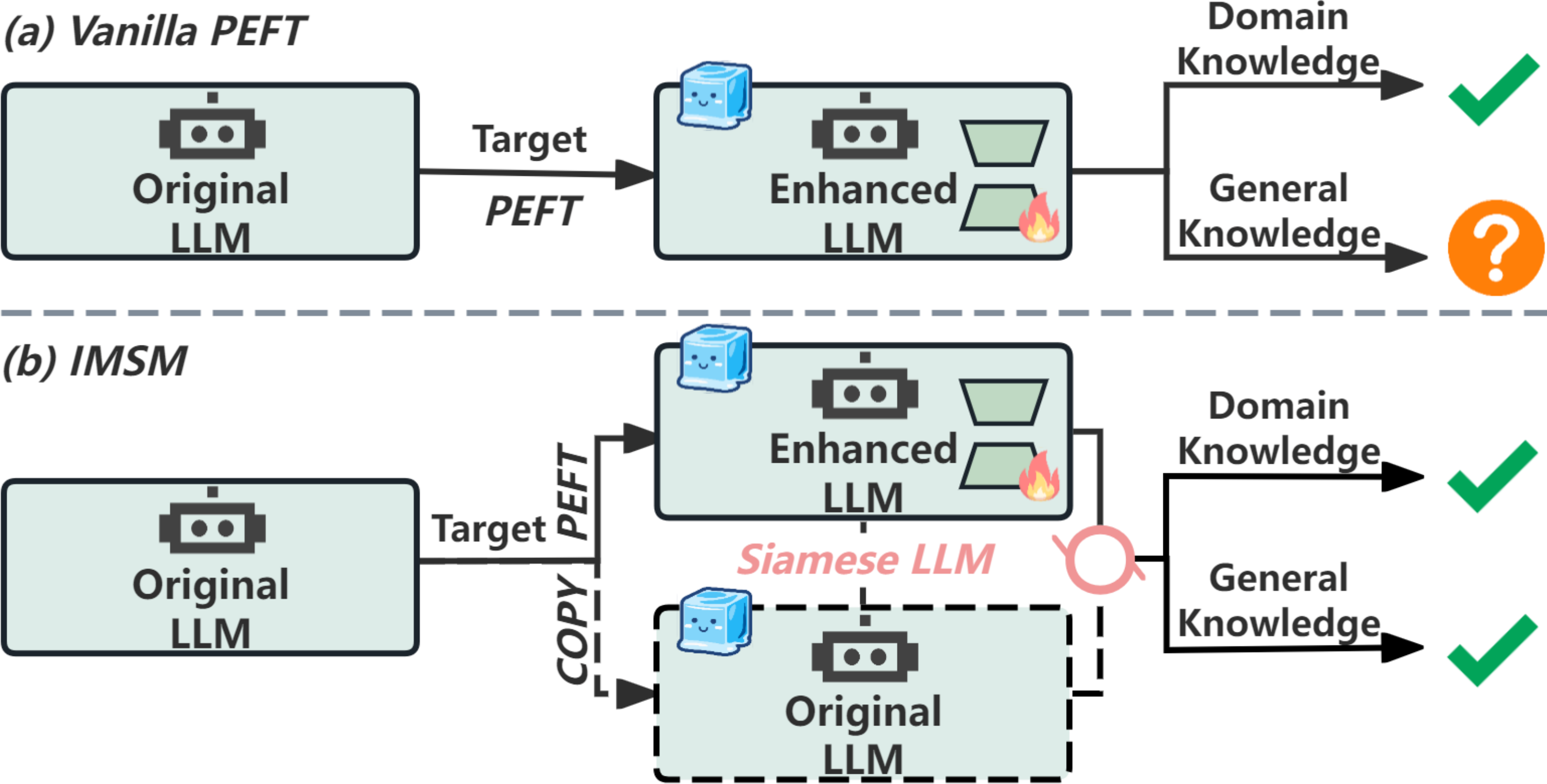}  
	\caption{Comparison between vanilla PEFT (a) and our proposed IMSM (b). Vanilla PEFT methods employ an LLM only once. The parameter distribution shift will cause the LLM to forget general knowledge. Instead, IMSM incorporates a siamese LLM, which can be regarded as two LLMs, sharing identical structure and pre-trained parameters. One remains frozen while the other is fine-tuned using an existing PEFT method. By flexibly recalling the memory of the original LLM, IMSM can improve fine-tuning performance and alleviate catastrophic forgetting.}	

	\label{fig:mymodel_intro}
\end{figure}

Large Language Models (LLMs) have emerged as a significant breakthrough in natural language processing (NLP) \cite{bai2023qwen, touvron2023llama,openai2023gpt,zeng2022glm}. 
With extensive world knowledge, they have exhibited remarkable zero-shot abilities across various tasks, such as language understanding \cite{achiam2023gpt}, logical reasoning \cite{kojima2022large}, \textit{etc}. 
Nevertheless, one of the challenges with LLMs is their susceptibility to generating incorrect or unfaithful outputs, especially in unfamiliar domains \cite{ sun2024trustllm, lin2024recurrent}. 
In-context Learning (ICL) \cite{brown2020language,  huang2024text} mitigates this issue by incorporating typical examples into prompts, but does not inherently align LLMs with downstream objectives \cite{fu2023effectiveness,ding2023parameter}.
Continued pre-training or supervised fine-tuning can better align parameters with tasks but is time-consuming and computationally intensive \cite{vm2024fine,han2024parameter}.
To address this issue, parameter-efficient fine-tuning (PEFT) \cite{fu2023effectiveness,ding2023parameter}, as shown in Figure \ref{fig:mymodel_intro} (a), selectively fine-tunes a limited subset of parameters or adds new ones, while keeping the majority fixed. 

However, refining the parameters of LLMs using target datasets enhances domain-specific knowledge but introduces biases related to that domain and task \cite{ren2024analyzing}. This process can result in forgetting world knowledge and compromising general abilities, which is known as catastrophic forgetting \cite{mccloskey1989catastrophic, kirkpatrick2017overcoming}.
Effective countermeasures typically encompass strategies for replaying pre-training data \cite{hayes2020remind, kar2022preventing} or constraints on updating parameters \cite{chen2020recall}.
Nevertheless, obtaining appropriate pre-training data for LLMs presents considerable challenges. 
Additionally, the vast range of tasks and the sheer number of parameters complicate efforts to balance integrating new knowledge with preserving existing capabilities. This complexity introduces uncontrollable variables during the fine-tuning process.

Instead, this paper highlights the significance of utilizing the original knowledge of LLM to improve fine-tuning effectiveness and reduce the risk of catastrophic forgetting, as depicted in Figure \ref{fig:mymodel_intro} (b).
Generally, the hidden states of the final layer of an LLM can be conceptualized as a memory. It encapsulates the encoded input sequence along with the preceding tokens of the generated response.
When the model processes a query, the new memory shaped by the updated knowledge from fine-tuning may diverge from its original understanding. 
Therefore, for tasks beyond the fine-tuning target, the predicted probability distribution of the next token from this new memory runs the risk of deviating from the typically accurate distribution.
In such cases, recalling the original memory before predicting based on the new memory at each step can help prevent deviations in token predictions.
This process is akin to opening a channel between two parallel worlds with different experiences, facilitating the fusion of memories of diverse values.

To this end, we propose a straightforward yet highly effective PEFT framework called \textbf{IMSM}, which \textbf{I}nterweaves \textbf{M}emories of a \textbf{S}iamese Large Language \textbf{M}odel. 
In particular, we employ a siamese LLM equipped with an existing PEFT method, like LoRA. Given a query, it generates two distinct final hidden states, \textit{i.e.}, memories, based on the original pre-trained parameters and the parameters equipped with the PEFT module. We then propose a query-aware gate as the memory interweaving mechanism to facilitate token generation at each step. 
During the training phase, the existing PEFT module effectively retains new knowledge for downstream tasks. Thus, the gating mechanism is capable of dynamically balancing between the original memory and the updated memory based on varying queries.
During the inference stage, IMSM relies on the given query features, allowing it to flexibly meet the demands of original knowledge for both fine-tuned tasks and other tasks. This mechanism is key to improving performance on fine-tuning tasks and reducing catastrophic forgetting on other tasks.
Theoretically, our proposed IMSM is a model-agnostic framework that allows for seamless integration across various open-source LLMs and existing PEFT methods. 

We conduct extensive experiments to evaluate the performance of the proposed IMSM in comparison to classical PEFT methods, including LoRA \cite{hu2021lora}, $(IA)^3$ \cite{liu2022few}, and AdaLoRA \cite{zhang2023adaptive}. We also compare with the state-of-the-art PEFT methods, LoRAMoE \cite{dou2024loramoe} and DoRA \cite{liudora}. Moreover, we employ popular LLMs such as ChatGLM3 \cite{du2021glm}, Qwen1.5 \cite{bai2023qwen}, Llama2 \cite{touvron2023llama}, and Llama3 \cite{touvron2023llama} as backbone LLMs and fine-tune them with PEFT on four benchmark datasets, \textit{e.g.}, MRPC \cite{el2015exploiting}, CoLA \cite{el2015exploiting}, ROPES \cite{lin2019reasoning}, and GSM8K \cite{cobbe2021training}. 
To evaluate the extent of catastrophic forgetting, we analyze the LLMs' performance on other datasets before and after fine-tuning on a target dataset.
The results demonstrate that IMSM achieves superior performance compared with vanilla PEFT methods.
Notably, this improvement is attained without imposing too many additional trainable parameters.
Furthermore, IMSM can effectively mitigate the issue of catastrophic forgetting.

Our main contributions are summarized as follows:

$\bullet$ To the best of our knowledge, this represents the first model-agnostic PEFT framework that employs a siamese LLM, effectively harnessing both its intrinsic world knowledge and the novel insights gained from PEFT.

$\bullet$ We propose a query-aware gate mechanism for interweaving memories, allowing flexible fusion of the last hidden states at each step.

$\bullet$ Extensive experiments validate the effectiveness of our proposed IMSM, and the results have confirmed its exceptional performance in striking a balance between plasticity and stability.

\section{Related Work}
\label{sec:relate}

We review three lines of related work: parameter-efficient fine-tuning methods, strategies for mitigating catastrophic in LLMs, and approaches employing logits arithmetic. 

\subsection{Parameter-efficient Fine-Tuning}
Traditional approaches to fully fine-tuning LLMs with billions of parameters suffer from significant difficulties in terms of training time, computational expense, and practical efficiency. To overcome these, parameter-efficient fine-tuning (PEFT) techniques have been developed, which focus on adjusting only a subset of the model weights while maintaining the rest unchanged \cite{mao2021unipelt}. 

PEFT methods can be categorized into four main types: additive, selective, re-parameterized, and hybrid approaches \cite{han2024parameter}. Additive methods, like prefix-tuning \cite{li2021prefix} and $(IA)^3$ \cite{liu2022few}, introduce additional modules into the LLM layers. Selective methods, like BitFit \cite{zaken2022bitfit}, focus on selecting a small subset of parameters for fine-tuning. Re-parameterized methods \cite{hu2021lora} use low-rank matrices to approximate the changing weights. Hybrid methods \cite{mao2022unipelt} are designed to combine different PEFT methods, offering a comprehensive solution. 

Current PEFT methods enable domain alignment at the parameter level but can risk disrupting LLMs' existing world knowledge and reasoning capabilities. Therefore, we propose IMSM, a model-agnostic framework for any open-source LLM and PEFT, which leverages the knowledge of the original LLM to enhance response accuracy by integrating the original memories generated from the same input.

\subsection{Catastrophic Forgetting in LLMs}

LLMs face the challenge of catastrophic forgetting \cite{mccloskey1989catastrophic} during both pre-training and fine-tuning, where the model tends to forget previously acquired knowledge after being fine-tuned for downstream tasks.
Catastrophic forgetting is the traditional challenge of continuous learning, leading to the development of various methods to address this issue in this scenario.
Prominent strategies include techniques such as constrained parameter updates, data replay, and parameter isolation \cite{ke2022continual}. 

Recently, empirical investigations have indicated that even advanced fine-tuning techniques like LoRA are not immune to catastrophic forgetting \cite{luo2023empirical}.  
To address the issue during PEFT, acquiring extensive pre-training data can be an expensive process. 
Moreover, it is challenging to ensure that the existing knowledge learned by the LLM is not overwritten at the parameter level. Correspondingly, this work suggests a novel approach to address this issue by utilizing the siamese model to directly recall the knowledge from the original LLM. By interweaving distinct memories for the same input, the responses can be generated with enhanced flexibility.

\subsection{Logits Arithmetic}
Due to the limitations of individual LLM, model collaboration emerges as a promising prospect. Previous studies \cite{liu2021dexperts} have highlighted the effectiveness of ensembling logits from multiple LLMs in controlled text generation or reducing hallucinations. 
Contrastive decoding \cite{o2023contrastive} is proposed to directly use the discrepancy in logits between robust and weaker models. Prefix-Adaptive Decoding \cite{pei2023preadd} and Context-aware decoding \cite{shi2023trusting} enhance the model's faithfulness by amplifying output probability differences across prompts. 

Logits are derived from the final hidden states of LLMs, which serve as internal representations or memories. Prior research \cite{azaria2023internal,duan2024llms} has shown that these hidden states encapsulate knowledge or information pertinent to factual judgments. Thus, the final hidden states, focused on generating the next token, reflect the LLMs' comprehension of the input information.

Taking this as a starting point, and different from previous methods, we place the siamese LLM within the PEFT framework, enabling a learnable collaboration. This approach not only enhances the performance of PEFT in downstream tasks, but also retains extensive world knowledge and reasoning capabilities of the original LLM. 

\section{Methodology}
\label{sec:model}

\begin{figure}[h]
	\centering
	\includegraphics[width=0.32\textwidth]{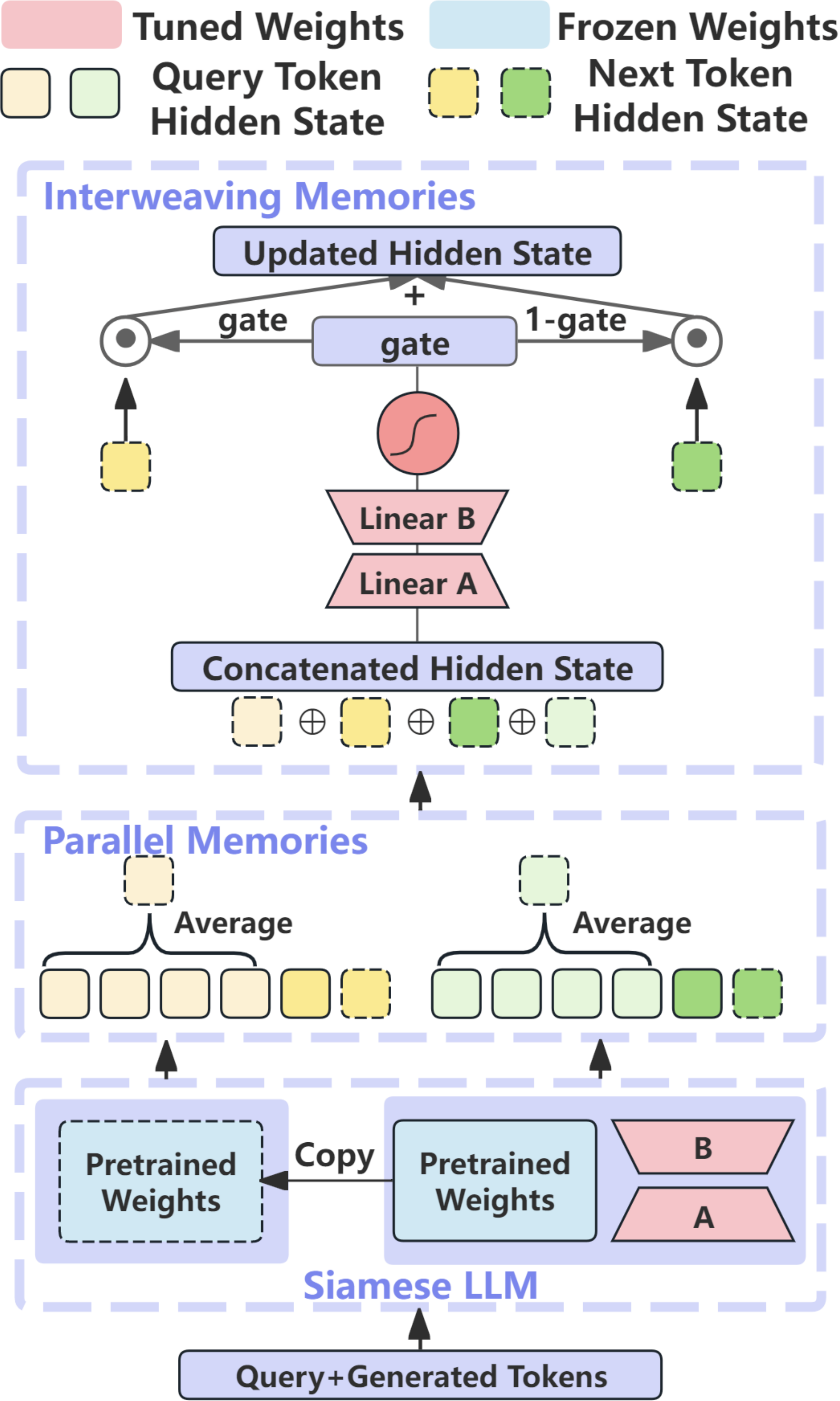}
	\caption{The overall architecture of IMSM, including a siamese LLM and an interweaving mechanism. Given the same input tokens, our siamese LLM produces memories with distinct values, which correspond to the two last hidden states. The generation of the next token relies on the updated memory through an interweaving mechanism. Trainable parameters are marked in red.}
	\label{fig:mymode_method}
\end{figure}

In this section, we introduce the proposed IMSM, a novel model-agnostic PEFT framework. As illustrated in Figure \ref{fig:mymode_method}, the siamese LLM is enhanced with a PEFT module. We conceptualize this siamese LLM as comprising two LLMs: one retains the original pre-trained knowledge, while the other is adapted with knowledge specific to the downstream task.
The input tokens and the previously generated response tokens are processed by the siamese LLM to create distinct memories, specifically the final hidden states. We then propose a query-aware gate mechanism to interweave these memory representations. The next token is generated from the intertwined memory.

In this framework, the parameters associated with PEFT and the parameters related to the gate mechanism will undergo fine-tuning, while the remaining parameters will be frozen. 
Our objective is to preserve the model's inherent world knowledge and general reasoning capabilities while simultaneously improving its performance in both downstream and general tasks. IMSM has the potential to be applied to various open-source LLMs and PEFT methods.

\subsection{Siamese Large Language Model}
To facilitate description and enhance intuitive understanding, our Siamese LLM can be conceptualized as a LLM with dual types of parameters.
When integrating PEFT methods into the siamese LLM, the model first retains a copy of the original knowledge. It then undergoes fine-tuning through the PEFT module using downstream data to better align with the new target objective. Taking LoRA as an exemplar PEFT method, we update the parameters $\bm{W}$ by adding an adapter $\bm{\Delta W}$ to the original parameters $\bm{W_o} \in \bm{\mathbb{R}}^{d \times k}$:
\begin{equation}
	\bm{W} = \bm{W}_{0} + \bm{\Delta W} = \bm{W}_{0} + \bm{B}\bm{A}
	\label{eq:LoRA}
\end{equation}
where $\bm{W_o}$ remains frozen, while $\bm{B} \in \bm{\mathbb{R}}^{d \times r'}$ and $\bm{A} \in \bm{\mathbb{R}}^{r' \times k}$ denote trainable low-rank matrices. And $r'$ is significantly smaller than $d$ and $k$. This configuration results in a substantial reduction in the number of trainable parameters. $\bm{W_o}$ and $\bm{W}$ represent the parameters of the frozen LLM (\textit{i.e.}, original LLM) and tuned LLM, respectively.

\subsection{Memory Interweaving Mechanism}

Throughout both the fine-tuning and inference processes, our siamese LLM generates distinct hidden states and corresponding logits. 
The last hidden states can serve as feature representations, encapsulating the LLM's memory of the previous input tokens.
We shall employ these two representations as vehicles to interweave memories derived from various experiences, and propose a gate-based interweaving mechanism for memories.

Let $\mathcal{M} $ represent the original LLM with parameters $\bm{W}_o$, while $\mathcal{M}^{'} $ denotes the fine-tuned LLM with parameters $\bm{W}$ including partially tuned parameters $\bm{\Delta W}$. 

At each time step ${t}$, we can access the hidden states derived from $\mathcal{M} $ and $\mathcal{M}^{'} $, as well as the hidden states prior to ${t}$. First, we compute the average of the query's last layer hidden states in the Siamese LLM to obtain its dense representations, reflecting the model's dual understanding of the input. Next, the hidden states for query understanding and next token prediction are concatenated along the feature dimension. Finally, we construct a linear layer with low-rank weight matrices at the top of the model to generate a gate, which determines the contribution of each LLM in the next step of logits generation process: 
\begin{equation}
\overline{\bm{h}}_{\mathcal{M}}^{q} = \frac{1}{T_{in}} \sum_{t=1}^{T_{in}} \bm{h}_{\mathcal{M}}^{t}
\end{equation}
\begin{equation}
\overline{\bm{h}}_{\mathcal{M}'}^{q} = \frac{1}{T_{in}} \sum_{t=1}^{T_{in}} \bm{h}_{\mathcal{M}'}^{t}
\end{equation}
\begin{equation}
	\begin{split}
		\bm{gate} &= \text{sigmoid}\left(f(\overline{\bm{h}}_{\mathcal{M}}^{q} \oplus \bm{h}_{\mathcal{M}}^t \oplus \bm{h}_{\mathcal{M}'}^t \oplus \overline{\bm{h}}_{\mathcal{M}'}^{q})\right) \\
		&= \text{sigmoid}\left((\overline{\bm{h}}_{\mathcal{M}}^{q} \oplus \bm{h}_{\mathcal{M}} \oplus \bm{h}_{\mathcal{M}'} \oplus \overline{\bm{h}}_{\mathcal{M}'}^{q})\cdot\bm{W}_A\cdot\bm{W}_B\right)
	\end{split}
\end{equation}
where $\bm{{h}}_{\mathcal{M}}^t$ and $\bm{h}_{\mathcal{M}'}^t$ represent the last layer hidden states of the original LLM $\mathcal{M}$ and the tuned LLM $\mathcal{M}^{'}$, respectively, for the $t$-th token in the sequence. $T_{in}$ represents the input sequence length. $\overline{\bm{h}}_{\mathcal{M}}^{q}$ and $\overline{\bm{h}}_{\mathcal{M}'}^{q}$ represent the average of the query’s last layer hidden states. $\oplus$ denotes vector concatenation and $f$ represents the linear layer with low rank. Here, $\bm{{h}}^t_{\mathcal{M}}, \bm{h}^t_{\mathcal{M}'} \in \mathbb{R}^{1 \times d}$, where $d$ represents the dimension of the hidden state. $\bm{{W}}_A \in \mathbb{R}^{4d \times r}$ and $\bm{{W}}_B \in \mathbb{R}^{r \times d}$, where $r$ is a hyper-parameter. Notably, $r$ is significant smaller than $d$, ensuring a limited number of parameters.
Furthermore, the gate is obtained through sigmoid activation. 

The updated last hidden state $\bm{h}^l_{\mathcal{N}}$ and the logits can be calculated by:

\begin{equation}
	\bm{h}^t_{\mathcal{N}} = \bm{gate} \circ \bm{h}^t_{\mathcal{M}} + (1 - \bm{gate}) \circ \bm{h}^t_{\mathcal{M}'} \quad 
\end{equation}
\begin{equation}
	\bm{logits} = \bm{h}^t_{\mathcal{N}}\cdot \bm{W}_{out}
\end{equation}
where $\circ$ denotes element-wise multiplication and $\bm{W}_{out}$ denotes the parameters of the linear layer of the siamese LLM that are utilized for generating the logits of the next token corresponding to the vocabulary size. 

In the memory interweaving process, incorporating query information into the gating mechanism enables the model to better retain original knowledge and reasoning abilities. Unlike additive methods that lead to parameter sharing, concatenating hidden states facilitates finer-grained feature fusion. Generally, both averaging and maximization are valid strategies for handling the parallel memories of past query tokens. However, averaging to synthesize information offers a more comprehensive representation of the query.
 
\begin{table*}[!h] % htbp
  \centering
    \begin{tabular}{c|c|c|cc|c|cc|c|c}
    \toprule
    \multirow{2}[4]{*}{\textbf{Backbone}} & \multirow{2}[4]{*}{\textbf{Method}} & \multirow{2}[4]{*}{\textbf{Params}} & \multicolumn{2}{c|}{\textbf{MRPC}} & \textbf{CoLA} & \multicolumn{2}{c|}{\textbf{ROPES}} & \textbf{GSM8K} & \multirow{2}[4]{*}{\textbf{Average}} \\
    \cmidrule{4-9}          &       &       & \textbf{Acc.} & \textbf{F1} & \textbf{Mcc.} & \textbf{EM} & \textbf{F1} & \textbf{Acc.} &  \\
    \midrule
    \multirow{7}[7]{*}{ChatGLM3} & Original & -     & 74.72  & 82.88  & 43.99  & 58.47  & 22.67  & 45.79  & 54.75  \\
    \cmidrule{2-10}          & LoRA  & 3.899M & 86.67  & 90.10  & 63.56  & 74.38  & 79.41  & 44.50  & 73.10  \\
          & IMSM  & 4.063M & \textbf{87.13*} & \textbf{90.49*} & \textbf{65.21*} & \textbf{79.15*} & \textbf{81.28*} & \textbf{48.45*} & \textbf{75.29*} \\
    \cmidrule{2-10}          & {${(IA)^3}$}   & 0.513M & 86.96  & 90.32  & 62.12  & 72.22  & 76.36  & 50.04  & 73.00  \\
          & IMSM  & 0.676M & \textbf{88.70*} & \textbf{91.66*} & \textbf{64.53*} & \textbf{73.34*} & \textbf{77.56*} & \textbf{52.46*} & \textbf{74.71*} \\
    \cmidrule{2-10}          & AdaLoRA & 2.925M & 87.02  & 90.35  & 62.48  & 75.14  & 78.64  & 49.36  & 73.83  \\
          & IMSM  & 3.089M & \textbf{87.77*} & \textbf{90.81*} & \textbf{65.57*} & \textbf{78.50*} & \textbf{82.23*} & \textbf{50.80*} & \textbf{75.95*} \\
    \midrule
    \multirow{7}[7]{*}{Qwen1.5} & Original & -     & 74.20  & 81.04  & 46.50  & 56.58  & 38.58  & 18.35  & 52.54  \\
    \cmidrule{2-10}          & LoRA  & 3.277M & 86.41  & 89.79  & 65.23  & 63.63  & 69.56  & 50.27  & 70.82  \\
          & IMSM  & 3.379M & \textbf{87.25*} & \textbf{90.48*} & \textbf{65.80*} & \textbf{65.64*} & \textbf{72.71*} & \textbf{52.69*} & \textbf{72.43*} \\
    \cmidrule{2-10}          & {${(IA)^3}$}   & 0.205M & 86.84  & 90.16  & 64.96  & 56.46  & 64.35  & 51.33  & 69.02  \\
          & IMSM  & 0.307M & \textbf{87.13*} & \textbf{90.73*} & \textbf{69.13*} & \textbf{60.25*} & \textbf{67.62*} & \textbf{53.15*} & \textbf{71.34*} \\
    \cmidrule{2-10}          & AdaLoRA & 2.458M & 86.73  & 90.15  & 65.63  & 64.81  & 68.62  & \textbf{52.01 } & 71.33  \\
          & IMSM  & 2.560M & \textbf{88.41*} & \textbf{91.32*} & \textbf{66.24*} & \textbf{67.18*} & \textbf{72.36*} & 51.86  & \textbf{72.90*} \\
    \midrule
    \multirow{7}[8]{*}{Llama3} & Original & -     & 68.17  & 79.83  & 37.96  & 49.23  & 13.95  & 3.23  & 42.06  \\
    \cmidrule{2-10}          & LoRA  & 6.816M & \textbf{89.10 } & \textbf{91.88} & 71.65  & 87.26  & 89.09  & 59.59  & 81.43  \\
          & IMSM  & 6.980M & 89.04  & 91.84  & \textbf{72.05*} & \textbf{87.85*} & \textbf{90.42*} & \textbf{61.26*} & \textbf{82.08*} \\
    \cmidrule{2-10}          & {${(IA)^3}$}   & 0.524M & 87.71  & 90.78  & 68.16  & 78.55  & 82.86  & \textbf{62.62 } & 78.45  \\
          & IMSM  & 0.688M & \textbf{88.87*} & \textbf{91.77*} & \textbf{70.78*} & \textbf{81.58*} & \textbf{85.95*} & 61.33  & \textbf{80.05*} \\
    \cmidrule{2-10}          & AdaLoRA & 5.113M & 88.87  & 91.72  & 68.77  & 87.21  & 88.98  & 55.42  & 80.16  \\
          & IMSM  & 5.276M & \textbf{89.22*} & \textbf{91.95*} & \textbf{71.55*} & \textbf{87.80*} & \textbf{89.05*} & \textbf{59.74*} & \textbf{81.55*} \\
    \bottomrule
    \end{tabular}%
    \caption{The overall comparison across four downstream tasks. The best results achieved using IMSM and its corresponding vanilla fine-tuning methods are highlighted in boldface. The improvements achieved by IMSM over all baselines are statistically significant, as measured by student's t-test with a significance level of $p < 0.05$.} 
  \label{tab:performance}%
\end{table*}%

\subsection{Training and Inference}

The probability distribution of the next token can be calculated as follows: 
\begin{equation}
	p(y_t|x, y_{<t}) = \mathit{softmax}(\bm{logits})
\end{equation}

During the fine-tuning phase, we utilize cross-entropy as our loss function:
\begin{equation}
	\mathcal{L}({\bm{\Delta W}},\bm{W}_A,\bm{W}_B) = -\sum_{t=1}^{T_{out}} \sum_{y \in V} y_{t} \mathit{log} \, p(y_t|x, y_{<t}) 
\end{equation}
where $T_{out}$ is the length of the output sequence. Finally, we optimize the parameters ${\bm{\Delta W}}$, $\bm{W}_A$, and $\bm{W}_B$ using the optimization algorithm like AdamW \cite{Loshchilov2017DecoupledWD}. Consequently, the gate mechanism is capable of learning to execute flexible memory interweaving for diverse queries. This allows our proposed IMSM to mitigate the interference caused by extraneous new knowledge for tasks that fall outside the fine-tuning objective.

During the inference phase, as shown in Algorithm~\ref{alg:inference}, we utilize a greedy search strategy to generate the output sequence. We select the token with the highest probability, informed by the updated memory, as the next generated token, add it to the input sequence, and then repeat this process until the complete sequence is generated.

\section{Experiments}
\label{sec:exp}

\subsection{Datasets}
We conduct experiments on four datasets: MRPC \cite{el2015exploiting}, CoLA \cite{el2015exploiting}, ROPES \cite{lin2019reasoning}, and GSM8K \cite{cobbe2021training}, to evaluate the alignment capability of our IMSM. Following previous studies \cite{schick2024toolformer,asai2023self}, we also employ MRPC \cite{el2015exploiting}, WebQ \cite{berant2013semantic}, FreebaseQA \cite{jiang2019freebaseqa}, and MultiRC \cite{khashabi2018looking}, to assess the abilities to retain general knowledge of LLM.
Distinct metrics are employed for different tasks.
More details about datasets and corresponding metrics can be found in Appendix A2.

\subsection{Backbone LLMs}
We employ four mainstream open-source LLMs, including ChatGLM3-6B \cite{zeng2022glm}, Qwen1.5-4B \cite{bai2023qwen}, Llama2-7B \cite{touvron2023llama}, and Llama3-8B \cite{touvron2023llama}, as backbones of our siamese LLM. We also directly prompt these LLMs as baseline models. More details can be found in Appendix A2.

\subsection{Baseline PEFT Methods}

We utilize classical and state-of-the-art PEFT methods, like LoRA \cite{hu2021lora}, {$(IA)^3$} \cite{liu2022few}, AdaLoRA \cite{zhang2023adaptive}, and DoRA \cite{liudora}, to perform fine-tuning on the siamese LLM of our IMSM. In addition, we compare LoRAMoE \cite{dou2024loramoe}, a plugin-based mixture of experts model for preventing catastrophic forgetting. More details can be found in Appendix A2.

\subsection{Implementation Details}

We utilize HuggingFace Transformers \cite{wolf2019huggingface} and PEFT \cite{peft} to perform our experiments. The fine-tuning procedure is executed on 8 NVIDIA A800 GPUs under a Linux system.

For LoRA, AdaLoRA, and DoRA, we employ AdamW as the optimizer with learning rates of $3 \times 10^{-4}$, $2 \times 10^{-3}$, and $1 \times 10^{-4}$, respectively, and a batch size of 16. The rank and alpha for LoRA are 16, while for DoRA, they follow the recommended settings of 16 and 32. For $(IA)^3$, we use Adafactor with a learning rate of $3 \times 10^{-3}$ and a batch size of 8. All methods are trained for 3 epochs. For LoRAMoE, we use the original paper's configuration. 
For a fair comparison, we set the configurations of the tuned target modules of IMSM to be exactly the same as vanilla PEFT, as detailed in Appendix A3. The gate rank $r$ of IMSM is set to 8. 

\subsection{Performance Comparison}

Table \ref{tab:performance} presents the fine-tuning results of IMSM and baselines across four datasets. Generally, the fine-tuned LLMs outperform the solely prompted LLMs. This highlights the significance of parameter tuning compared with ICL for aligning LLMs to unfamiliar tasks. Moreover, our proposed method, IMSM, significantly surpasses all vanilla PEFT methods across all three LLMs.

The effect of parameter fine-tuning is influenced by multiple factors, including the quality and capabilities of the backbone LLM, and the adaptability of the employed PEFT method.
As one of the most advanced open-source LLMs, the Llama3-based models achieve near-optimal performance. Furthermore, AdaLoRA dynamically adjusts the rank of low-rank matrices, highlighting its efficacy in aligning with downstream tasks.

IMSM, as a simple yet powerful model-agnostic PEFT framework, excels at integrating the stability of the original LLM with the adaptability of the tuned LLM. Additionally, it incorporates the gate mechanism that allows for selective control over whether the understanding is derived from the fixed or adjusted knowledge. Therefore, it leads to consistent improvement across various downstream tasks.

\begin{table}[] % htbp
  \centering
  \small 
  \setlength{\tabcolsep}{2.4pt} 
    \begin{tabular}{c|c|cccc|c}
    \toprule
    \textbf{Dataset} & \textbf{ROPES} & \textbf{WebQ} & \textbf{MultiRC} & \textbf{MRPC} & \textbf{Freebase} & \textbf{Avg.} \\
    \midrule
    ChatGLM3 & 58.47 & 25.49 & 71.74 & 74.72 & 36.81 & 53.45  \\
    \midrule
    LoRA  & 74.38 & 20.47 & 66.67 & 70.32 & \textbf{38.61} & 54.09  \\
    IMSM  & \textbf{79.15} & \textbf{21.11} & \textbf{71.47} & \textbf{71.83} & 38.01 & \textbf{56.31} \\
    \midrule
    {${(IA)^3}$}   & 72.22 & 20.96 & 70.83 & 72.58 & 37.34 & 54.79  \\
    IMSM  & \textbf{73.34} & \textbf{21.36} & \textbf{73.72} & \textbf{74.20} & \textbf{37.74} & \textbf{56.07} \\
    \midrule
    AdaLoRA & 75.14 & 19.98 & 67.63 & 70.20  & 35.99 & 53.79  \\
    IMSM  & \textbf{78.50} & \textbf{22.88} & \textbf{73.72} & \textbf{74.96} & \textbf{38.41} & \textbf{57.69} \\
    \bottomrule
    \end{tabular}%
    \caption{
Catastrophic forgetting validation between vanilla PEFT and IMSM using ChatGLM3-6B, fine-tuned on ROPES and evaluated on general knowledge benchmarks.} %  ($p<0.05$)
  \label{tab:cf_on_ropes}%
\end{table}%

\begin{table}[htbp]
  \centering
  \small 
  \setlength{\tabcolsep}{2.3pt} 
  
    \begin{tabular}{c|c|cccc|c}
    \toprule
    \textbf{Dataset} & \textbf{GSM8K} & \textbf{WebQ} & \textbf{MultiRC} & \textbf{MRPC} & \textbf{Freebase} & \textbf{Avg.} \\
    \midrule
    ChatGLM3 & 45.79  & 25.49 & 71.74 & 74.72 & 36.81 & 50.91  \\
    \midrule
    DoRA  & 45.34  & \textbf{26.18} & 71.47 & 74.55  & 33.26  & 50.16  \\
    IMSM  & \textbf{48.75} & 25.49  & \textbf{75.00}  & \textbf{75.19} & \textbf{38.89} & \textbf{52.66} \\
    \midrule
    Qwen1.5 & 18.35  & 32.33 & 66.99 & 74.20  & 47.12 & 47.80  \\
    \midrule
    DoRA  & 51.33  & 27.51  & 66.03  & 73.91  & 41.19  & 51.99  \\
    IMSM  & \textbf{52.84} & \textbf{28.94} & \textbf{68.27} & \textbf{74.26} & \textbf{43.52} & \textbf{53.57} \\
    \midrule
    Llama3.1 & 33.06 & 41.78 & 54.17 & 42.43  & 76.35 & 49.56  \\
    \midrule
    DoRA  & 72.71  & 32.78  & 58.65  & \textbf{55.55} & 65.92  & 57.12  \\
    IMSM  & \textbf{73.77} & \textbf{40.31} & \textbf{66.90} & 47.94  & \textbf{75.45} & \textbf{60.87} \\
    \bottomrule
    \end{tabular}%
    \caption{Evaluation of catastrophic forgetting between DoRA and its corresponding IMSM across different LLMs.}
  \label{tab:dora_cf}%
\end{table}%

\subsection{Catastrophic Forgetting Evaluation} 

Fine-tuning inevitably introduces a degree of forgetting. We fine-tune LLMs on one dataset and subsequently evaluate their performance on other general benchmarks.

Table \ref{tab:cf_on_ropes} shows the results of fine-tuning ChatGLM on ROPES. For the three vanilla PEFT methods, they all prioritize enhancing adaptability while neglecting stability. After fine-tuning ChatGLM on ROPES with LoRA, $(IA)^3$, and AdaLoRA, the accuracy on MultiRC drops from 71.74 to 66.67, 70.83, and 67.63, respectively.  
IMSM can effectively mitigate performance degradation. For example, IMSM improves overall performance by 4.10\% compared with LoRA on ChatGLM. Notably, the interweaving strategy of IMSM can even revive precise memories within the LLM, enhancing performance on non-target datasets.

\begin{figure}
        \centering
	
\includegraphics[width=0.375\textwidth]{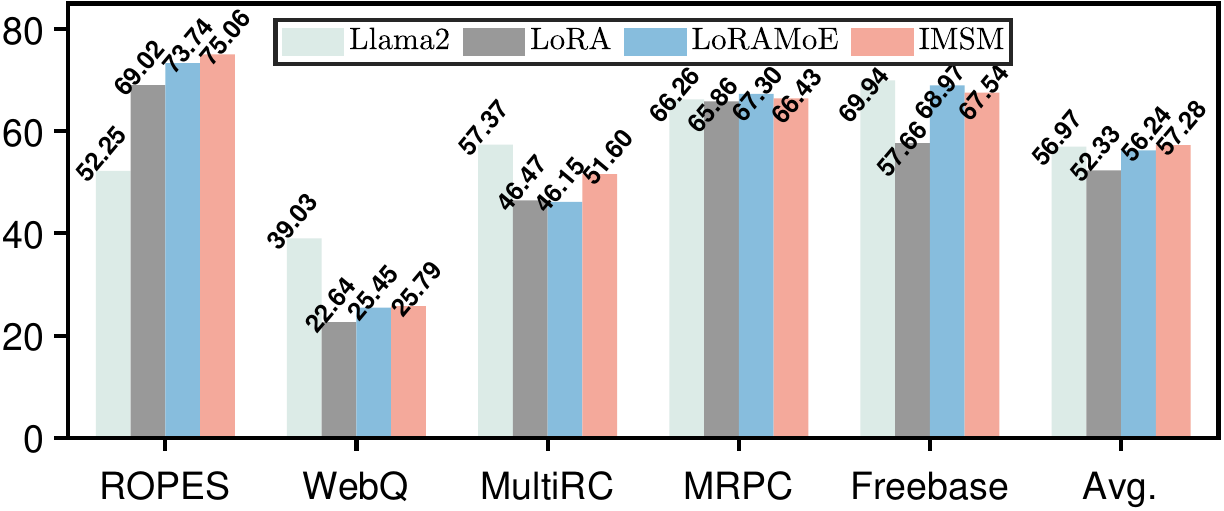} \includegraphics[width=0.375\textwidth]{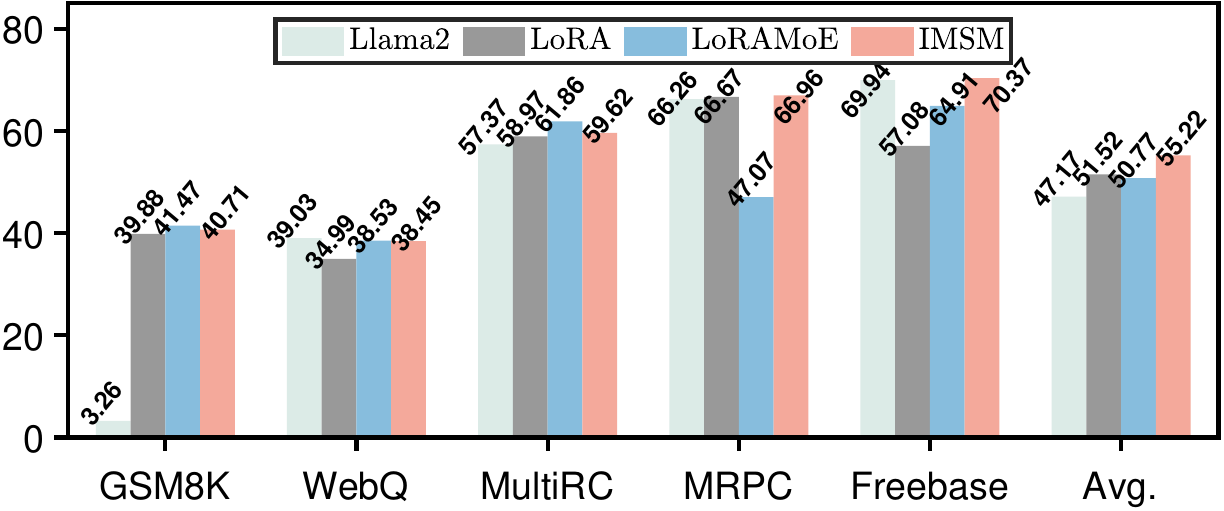}
	
	\caption{Evaluation of catastrophic forgetting using Llama2-7B, fine-tuned on ROPES and GSM8K, and evaluated on general knowledge benchmarks.}

	\label{fig:CF-LoRA-LoRAMoE_GSM8K}
\end{figure}

We also compare IMSM with LoRAMoE and DoRA, which are state-of-the-art solutions for catastrophic forgetting and PEFT, respectively. 
For LoRAMoE, we follow the original setup using Llama2 as the backbone. Due to crashes when using DoRA for fine-tuning Llama3, we opt for Llama3.1-instruct instead.
As illustrated in Figure \ref{fig:CF-LoRA-LoRAMoE_GSM8K}, IMSM reaches or surpasses LoRAMoE on general benchmarks, offering an average boost of 1.85\% and 8.77\% in overall performance. 
As shown in Table \ref{tab:dora_cf}, whether on the target dataset GSM8K or general benchmarks, IMSM's performance consistently outperforms the standalone DoRA.

The last hidden state serves as the memory for the LLMs' understanding of the input sequence. By considering the original and updated memories, we ensure that the siamese LLM retains a balance between its prior knowledge and its adaptation to new data. 

\begin{figure*}[!h]
\centering 
	\includegraphics[width=0.305\textwidth]{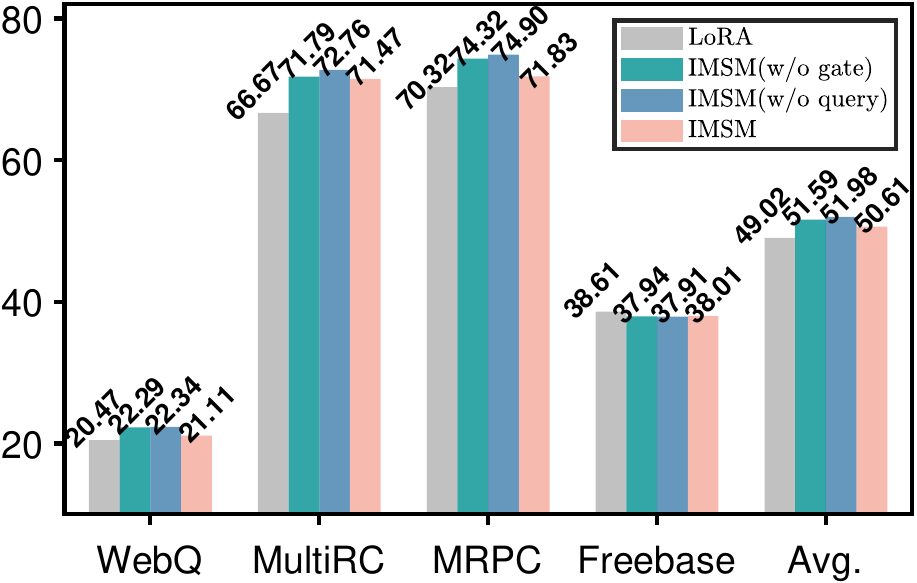}~ 
 \includegraphics[width=0.305\textwidth]{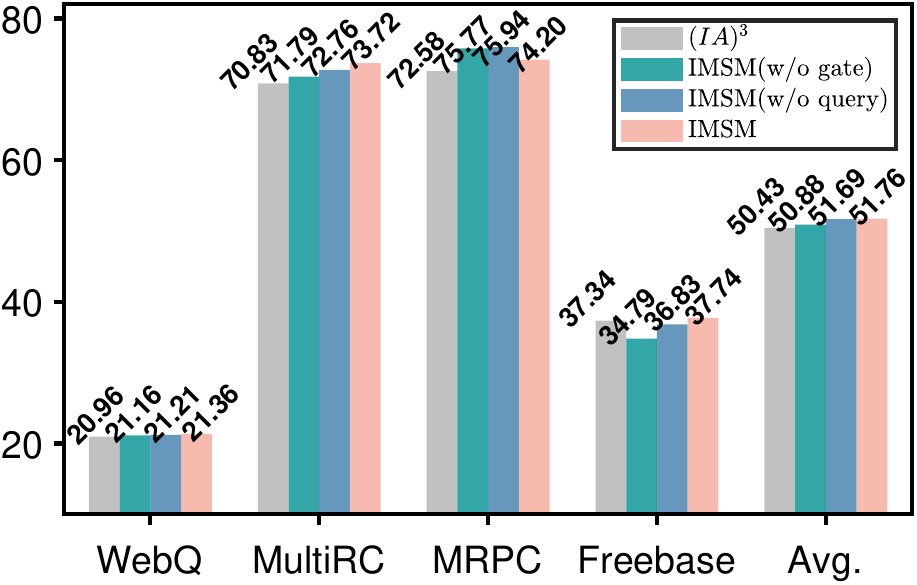}~ 
 \includegraphics[width=0.305\textwidth]{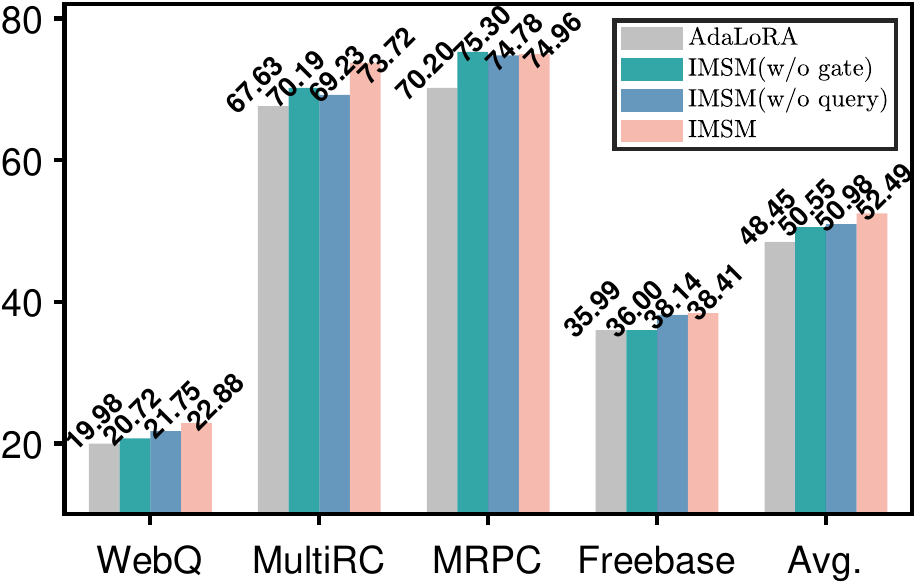}~ 
	
	\caption{Ablation test of catastrophic forgetting validation. ChatGLM3-6B is employed as the backbone LLM, fine-tuned on ROPES, and evaluated on general knowledge benchmarks.}

	\label{fig:ablation_test}
\end{figure*}

\subsection{Ablation Study}

We conduct ablation tests to investigate how the query-aware gate, which serves as the memory interweaving mechanism, improves performance and assess its necessity. The results are shown in Table \ref{tab:ablation_test} and Figure \ref{fig:ablation_test}. In the ``w/o query" setting, the siamese LLM's understanding of the query is excluded from the gate construction. In the ``w/o gate" setting, the gate construction is omitted, and the two final hidden states of the siamese LLM are directly added in a 1:1 ratio.

As shown in Table \ref{tab:ablation_test}, the query-aware gate mechanism brings the most significant improvement to downstream task performance in almost all settings. Figure \ref{fig:ablation_test} illustrates that directly combining memories and incorporating the original parameter memory bring performance gains on non-target datasets compared with vanilla PEFT. This highlights the effectiveness of recalling original knowledge through the final layer hidden state. With the introduction of a simple gate, the approach can dynamically balance the previous and the refreshed knowledge. The integration of the siamese LLM's query feature further alleviates catastrophic forgetting.

\begin{table}[htbp]
  \centering
  \setlength{\tabcolsep}{4.5pt} 
    \begin{tabular}{r|ccccc|c}
    \toprule
    \multicolumn{1}{c|}{\textbf{Dataset}} & \multicolumn{2}{c}{\textbf{MRPC}} & \textbf{CoLA} & \multicolumn{2}{c|}{\textbf{ROPES}} & \textbf{Avg.} \\
    \toprule
    \multicolumn{1}{c|}{LoRA} & 86.67 & 90.10  & 63.56 & 74.38 & 79.41 & 78.82  \\
    \midrule
    \multicolumn{1}{c|}{IMSM} & 87.13 & 90.49 & \textbf{65.21} & \textbf{79.15} & \textbf{81.28} & \textbf{80.65} \\
    w/o query & \textbf{87.62} & \textbf{90.81} & 64.92 & 77.43 & 80.38 & 80.23  \\
    w/o gate & 87.01 & 90.31 & 63.74 & 76.54 & 80.12 & 79.54  \\
    \toprule
    \multicolumn{1}{c|}{${(IA)^3}$} & 86.96 & 90.32 & 62.12 & 72.22 & 76.36 & 77.60  \\
    \midrule
    \multicolumn{1}{c|}{IMSM} & \textbf{88.70} & \textbf{91.66} & \textbf{64.53} & \textbf{73.34} & \textbf{77.56} & \textbf{79.16} \\
    w/o query & 87.77 & 91.04 & 63.49 & 72.93 & 76.73 & 78.39  \\
    w/o gate & 87.04 & 90.48 & 62.94 & 71.86 & 76.77 & 77.82  \\
    \toprule
    \multicolumn{1}{c|}{AdaLoRA} & 87.02 & 90.35 & 62.48 & 75.14 & 78.64 & 78.73  \\
    \midrule
    \multicolumn{1}{c|}{IMSM} & \textbf{87.77} & \textbf{90.81} & \textbf{65.57} & \textbf{78.50} & \textbf{82.23} & \textbf{80.98} \\
    w/o query & 87.59 & 90.68 & 65.19 & 75.41 & 79.56 & 79.69  \\
    w/o gate & 87.30  & 90.59 & 64.94 & 76.60  & 80.07 & 79.90  \\
    \bottomrule
    \end{tabular}%
    \caption{The ablation test is performed on ChatGLM3-6B to compare its performance across various downstream tasks.} 
  \label{tab:ablation_test}%
\end{table}%

\subsection{Space and Time Complexity}
Compared to the PEFT method alone, IMSM maintains comparable space and time complexity. While the gate mechanism introduces additional trainable parameters, their impact is minimal compared with the original PEFT. We report the count of trainable parameters in Table \ref{tab:performance}, with a gate rank of 8 used across all IMSM experiments.
During inference, the time required to interweave the two memories remains constant for each step, with the two memory streams being generated in parallel theoretically. Compared to vanilla PEFT, IMSM introduces only a lightweight gating mechanism overhead, maintaining equivalent time complexity. The inference speed, measured in tokens per second, is presented in Table \ref{tab:inference_speed}. The extra memory usage is a necessary and acceptable trade-off to address catastrophic forgetting.

\begin{table}[htbp]
  \centering
  \small
  
    \begin{tabular}{cccc}
    \toprule
    \textbf{Method} & \textbf{ChatGLM3} & \textbf{Qwen1.5} & \textbf{Llama3} \\
    \midrule
    LoRA  & 31.30  & 22.86  & 22.98  \\
    IMSM  & 29.05  & 21.76  & 21.18  \\
    \midrule
    {$(IA)^3$}   & 33.86  & 23.51  & 22.81  \\
    IMSM  & 30.40  & 22.10  & 20.51  \\
    \midrule
    AdaLoRA & 30.85  & 21.68  & 21.49  \\
    IMSM  & 27.91  & 20.03  & 18.71  \\
    \bottomrule
    \end{tabular}%
    \caption{The comparisons of inference speed (tokens/sec) between vanilla PEFT methods and IMSM.}
  \label{tab:inference_speed}%
\end{table}%

\subsection{Hyper-parameter Analysis}

This section focuses on the sensitivity of the rank $r$, a hyper-parameter in the gate mechanism. Keeping other settings constant, we vary the rank among 4, 8, and 16. 
Please refer to Table \ref{tab:hyper-paramter table} for a comprehensive representation. 
While increasing the rank adds more trainable parameters, it does not necessarily improve performance on the target dataset. 
The optimal rank varies depending on the specific PEFT used. For IMSM based on LoRA and ${(IA)^3}$, a rank of 8 is often optimal, while for AdaLoRA, 16 is usually the most beneficial choice. 
However, as the rank increases, the gate mechanism may better recall original memories, potentially reducing forgetting on non-target datasets more effectively.

\begin{table}[htbp]
  \centering
  \small 
  \setlength{\tabcolsep}{2.3pt} 
  
    \begin{tabular}{c|c|c|c|cc}
    \toprule
    \textbf{IMSM Method} & \textbf{Rank} & \textbf{Params} & \textbf{GSM8K} & \textbf{WebQ} & \textbf{Freebase} \\
    \midrule
    \multirow{3}[2]{*}{LoRA} & 4     & 6.898M & 59.44  & 40.55  & 73.15  \\
          & 8     & 6.980M & \textbf{61.26} & \textbf{40.90} & 73.72  \\
          & 16    & 7.143M & 59.21  & 39.86  & \textbf{75.23} \\
    \midrule
    \multirow{3}[2]{*}{$(IA)^3$} & 4     & 0.606M & 58.98  & 37.80  & 74.60  \\
          & 8     & 0.688M & \textbf{61.33} & 38.58  & 74.32  \\
          & 16    & 0.852M & 61.26  & \textbf{38.98} & \textbf{75.90} \\
    \midrule
    \multirow{3}[2]{*}{AdaLoRA} & 4     & 5.194M & 59.67  & 41.63  & 73.97  \\
          & 8     & 5.276M & 59.74  & 40.94  & 74.28  \\
          & 16    & 5.440M & \textbf{60.50} & \textbf{42.72} & \textbf{75.45} \\
    \bottomrule
    \end{tabular}
    \caption{Efficiency on gate rank $r$ is fine-tuned on GSM8K using Llama3-8B, with effects on WebQ and Freebase.}
  \label{tab:hyper-paramter table}
\end{table}

\section{Conclusion and Future Work}
\label{sec:conclu}

This paper presents a novel approach to fine-tuning LLMs, aiming to strike a delicate balance between plasticity and stability. 
Our proposed IMSM constitutes a model-agnostic framework applicable to any open-source LLMs, in conjunction with existing PEFT methods.
Particularly, IMSM enables the interweaving of memories derived from the siamese LLM, facilitating collaboration between frozen knowledge and tuned knowledge.
Extensive experiments substantiate that the interweaving mechanism significantly improves alignment performance on downstream tasks and alleviates catastrophic forgetting. 
Additional hyper-parameter experiments further confirm the robustness of the proposed model.

In future research, we aim to explore more intricate memory fusion mechanisms within multiple Transformer layers. Besides, we intend to evaluate the performance of our model on more challenging scenarios of catastrophic forgetting.

\section{Acknowledgments}
This work is supported by the National Natural Science Foundation of China (72204087, 72404212, 72234005), the Shanghai Planning Office of Philosophy and Social Science Youth Project (2022ETQ001), the “Chen Guang” project supported by Shanghai Municipal Education Commission and Shanghai Education Development Foundation (23CGA28), the Shanghai Pujiang Program (23PJC030), the Fundamental Research Funds for the Central Universities, China, and the 2024 Innovation Evaluation Open Fund, Fudan University (CXPJ2024006). We also appreciate the constructive comments from the anonymous reviewers.
\bibliography{arxiv}

\clearpage
\section{Appendix}
\subsection{A1. Inference Process}
\label{appendix:inferencce}

The inference process of the proposed IMSM is shown in Algorithm~\ref{alg:inference}.
\begin{algorithm}
	\caption{Inference Algorithm}
	\label{alg:inference}
	\begin{algorithmic}[1]
		
		\STATE \textbf{Input:} query $x$, query length $T_{in}$
		\STATE \textbf{Output:} output response $y$
		\STATE \textbf{Model:} the siamese LLM (can be regarded as a frozen LLM $\mathcal{M}$ and a tuned LLM $\mathcal{M}^{'}$), gate mechanism including $\bm{W}_A$ and $\bm{W}_B$, maximum output length $T$, EOS token $t_{\text{eos}}$, and vocabulary $\mathcal{V}$
		\STATE \textbf{Initialize:} $y \gets$ empty sequence
		
		\FOR{$t = 1$ to $T$}
		
             \IF{$t = 1$} 
             \STATE Calculate the siaseme LLM's understandings of the query and store them in the cache:
          
        \STATE \quad  $\overline{\bm{h}}_{\mathcal{M}}^{q} = \frac{1}{T_{in}} \sum_{t=1}^{T_{in}} \bm{h}_{\mathcal{M}}^{t}$

        \STATE \quad $\overline{\bm{h}}_{\mathcal{M}'}^{q} = \frac{1}{T_{in}} \sum_{t=1}^{T_{in}} \bm{h}_{\mathcal{M}'}^{t}$

	       \ENDIF
            \STATE Compute updated last layer hidden state based on $x$ and $y_{<t}$ using gate-based query-driven memory interweaving mechanism:

		\STATE \quad \quad $\bm{gate} = \text{sigmoid}(\left(\overline{\bm{h}}_{\mathcal{M}}^{q} \oplus \bm{h}^t_{\mathcal{M}} \oplus \bm{h}^t_{\mathcal{M}'} \oplus \overline{\bm{h}}_{\mathcal{M}'}^{q}\right) \cdot$ \\
\STATE \quad \quad \quad \quad $\bm{W}_A \cdot \bm{W}_B)$

		\STATE \quad \quad $\bm{h}^t_\mathcal{N} = \bm{gate} \circ \bm{h}^t_\mathcal{M}+(1-\bm{gate}) \circ \bm{h}^t_{\mathcal{M}^{'}}$
		\STATE Compute the next token probability distribution:
		% \STATE $\quad p(y_t^i|x, y_{<t}) = \text{softmax}(\bm{W}_{\text{out}}(\bm{h}^t_\mathcal{N}))$
  		\STATE $\quad p(y_t^i|x, y_{<t}) = \text{softmax}(\bm{h}^t_\mathcal{N}\cdot \bm{W}_{\text{out}})$

		\STATE Utilize greedy search to obtain the next token $y_{t}$
		\STATE Update the sequence: $x \mathrel{+}= y_{t}$, $y \mathrel{+}= y_{t}$
		\IF{$y_t = t_{\text{eos}}$} 
		\STATE break
		\ENDIF
		
		\ENDFOR
		
	\end{algorithmic}
\end{algorithm}

\subsection{A2. Evaluation Setup}
\label{appendix:evaluation_setup}

\subsubsection{Datasets}
The datasets used in this paper are as follows:
\begin{itemize}
    \item MRPC \cite{el2015exploiting} is a dataset that aims to determine whether two given sentences have the same meaning.
    \item CoLA \cite{el2015exploiting} is a linguistically annotated dataset designed to assess the model's ability to discern grammatical acceptability in sentences.
    \item ROPES \cite{lin2019reasoning} is a reading comprehension dataset designed to evaluate models' reasoning abilities demanding multi-step reasoning over a new situation.
    \item GSM8K \cite{cobbe2021training} is a collection of questions to evaluate the ability of LLM to solve mathematical word problems.
    \item WebQ \cite{berant2013semantic} is a question answering dataset to evaluate the world knowledge LLM.
    \item Freebase \cite{jiang2019freebaseqa} is also a context-free dataset used to evaluate the knowledge stored in LLM.
    \item MultiRC \cite{khashabi2018looking} is used to evaluate the model's understanding of short paragraphs and multi-sentence questions. We use a sub-dataset of the original dataset, same as the previous study \cite{dai2023long}.

\end{itemize}

\subsubsection{Metrics}
For MRPC, we typically use accuracy (Acc.) and F1 score as the metrics. For CoLA, we utilize mattews correlation coefficient (MCC) \cite{chicco2020advantages} as the primary metric. For ROPES, we employ Exact Match (EM) and F1 token overlap between the answer text and the golden truth as the metrics. For GSM8K and MultiRC, we typically use accuracy (Acc.) as the evaluation metric. Following \cite{schick2024toolformer,asai2023self}, we evaluate the performance on WebQ and Freebase by examining whether the generated response contains the golden answer rather than strictly requiring exact matching.

\subsubsection{Backbone LLMs}

We employ four mainstream open-source LLMs:
\begin{itemize}
	\item ChatGLM3-6B \cite{du2021glm,zeng2022glm}: Released in 2023, it optimizes positional encoding and utilizes the Swish activation function.
	\item Qwen1.5-4B  \cite{bai2023qwen}: With its release in 2024, it is built on the Transformer architecture with SwiGLU activation and attention QKV bias.
        \item Llama2-7B \cite{touvron2023llama}: Released in 2023, it improves upon Llama1 by increasing context length and introducing Grouped Query Attention (GQA).
	\item Llama3-8B \cite{touvron2023llama}: Also released in 2024, it employs an optimized transformer architecture, adopting Grouped Query Attention (GQA) to enhance inference efficiency.
\end{itemize}

\subsubsection{Baseline Methods}
We utilize the following methods as baselines: 
\begin{itemize}
    \item LoRA \cite{hu2021lora} freezes the pre-trained model weights and incorporates a detachable plugin, the trainable rank decomposition matrix, into the layers of the Transformer architecture. 
    \item {$(IA)^3$} \cite{liu2022few} rescales inner activations with learned vectors and requires fewer trainable parameters than LoRA.
    \item AdaLoRA \cite{zhang2023adaptive} adaptively adjusts the ranks of different modules based on the importance of the weight matrix.
    \item DoRA \cite{liudora} leverages finer-grained control by separately adjusting the magnitude and direction of the weights.
    \item LoRAMoE \cite{dou2024loramoe} is a plugin-based Mixture of Experts approach designed to mitigate knowledge forgetting in LLMs.
\end{itemize}

\subsection{A3. Implementation Details}
\label{appendix:target_modules}
The target adapters of LoRA \& AdaLoRA, ${(IA)^3}$, and DoRA are shown in Table \ref{tab:target_adapter_lora_adalora}, Table \ref{tab:target_adapter_ia3}, and Table \ref{tab:target_adapter_dora}, respectively.

\begin{table}[H]
  \centering
  \small
  \begin{tabular}{c|c}
    \toprule
    \textbf{Backbone} & \textbf{target\_modules} \\
    \midrule
    \textbf{ChatGLM3} & {[``query\_key\_value'']} \\
    \textbf{Qwen1.5}  & {[``q\_proj'']} \\
    \textbf{Llama3} & {[``q\_proj'', ``v\_proj'']} \\
    \textbf{Llama2} & {[``gate\_proj'', ``down\_proj'', ``up\_proj'']} \\
    \bottomrule
  \end{tabular}
  \caption{Target Adapter of LoRA \& AdaLoRA for Different Backbones.}
  \label{tab:target_adapter_lora_adalora}
\end{table}

\begin{table}[H]
  \centering
  \small
  \setlength{\tabcolsep}{3pt}  % Reduce column separation
  \begin{tabular}{c|c|c}
    \toprule
    \textbf{Backbone} & \parbox[c]{9.835em}{\centering \textbf{target\_modules}} & \textbf{feedforward\_modules} \\
    \midrule
    \textbf{ChatGLM3} & \parbox[c]{9.835em}{\centering[``query\_key\_value'',\\ ``mlp.dense\_4h\_to\_h'']} & [``mlp.dense\_4h\_to\_h''] \\
    \textbf{Qwen1.5}  & \parbox[c]{9.835em}{\centering[``q\_proj'', ``v\_proj'']} & [] \\
    \textbf{Llama3} & \parbox[c]{9.835em}{\centering [``k\_proj'', ``v\_proj'', ``down\_proj'']} & [``down\_proj''] \\
    \bottomrule
  \end{tabular}
  \caption{Target Adapter of {${(IA)^3}$} for Different Backbones.}
  \label{tab:target_adapter_ia3}
\end{table}

\begin{table}[H]
  \centering
  \small
  \begin{tabular}{c|c}
    \toprule
    \textbf{Backbone} & \textbf{target\_modules} \\
    \midrule
    \textbf{ChatGLM3} & [``query\_key\_value''] \\
    \textbf{Qwen1.5}  & [``q\_proj'', ``v\_proj''] \\
    \textbf{Llama3.1} & [``q\_proj'', ``v\_proj''] \\
    \bottomrule
  \end{tabular}
  \caption{Target Adapter of DoRA for Different Backbones.}
  \label{tab:target_adapter_dora}
\end{table}

\end{document}